\documentclass{article}

\usepackage{iclr2027_conference,times}

\usepackage{amsmath,amssymb,amsfonts,bm}
\usepackage{graphicx}
\usepackage{booktabs}
\usepackage{multirow}
\usepackage{array}
\usepackage[table]{xcolor}
\usepackage[colorlinks=true,linkcolor=black,citecolor=black,urlcolor=blue]{hyperref}
\iclrfinalcopy
\graphicspath{{figures/}}

\DeclareMathOperator*{\meanop}{mean}
\DeclareMathOperator{\stdop}{std}

\title{D-JEPA: Design-Recoverable JEPA Representation with Swappable Physics Decoders}

\author{%
\parbox{\textwidth}{%
\centering
Nitin Nagesh Kulkarni,
Aashwin Anand Mishra,
Yin Yu,
Peter Lyu\\
Luminary\\
San Mateo, CA, USA
}%
}

\begin{document}
\maketitle

\begin{abstract}
Joint-Embedding Predictive Architectures (JEPAs) provide a framework for
learning compact representations without directly reconstructing
high-dimensional observations. However, in parameterized physical systems,
learned representations can entangle geometry with operating conditions and
task-specific physical responses, limiting their reuse across prediction
tasks. We introduce D-JEPA (Design-recoverable JEPA), a geometry-centric JEPA
that computes a compact representation from geometry alone and reuses it
across operating conditions and physical response spaces through lightweight
physics-specific decoders. An explicit design-recoverability objective
encourages the geometry latent to preserve information about the underlying
design variables, enabling the representation to support design analysis and
optimization. We further identify a case-level collapse failure mode in which
target representations become nearly invariant across distinct geometries
despite low reconstruction error, and mitigate it using case-level variation
constraints and auxiliary target reconstruction. Across four 3D aerodynamic,
hydrodynamic, and structural benchmarks, D-JEPA maintains or improves
full-field prediction accuracy while achieving near-perfect linear
recoverability of design parameters. The frozen geometry representation can
be reused at held-out operating conditions and transferred to a structural
response task with fewer trainable parameters. Finally, the
representation supports differentiable design optimization, with
designs validated using high-fidelity CFD, preserving the predicted ranking
of candidate designs. These results demonstrate that separating a reusable
geometry representation from physics-specific prediction provides a
practical representation for scientific surrogate modeling and design.
\end{abstract}

\section{Introduction}

Many scientific and engineering problems require repeatedly solving parameterized partial differential equations (PDEs) as geometry, operating conditions, or boundary conditions vary. In aerodynamic design, Aerodynamic Databases (AeroDBs) map flight conditions and vehicle geometries to quantities of interest (QoIs) such as lift, drag, and structural loads, supporting applications from preliminary aircraft design and aeroelasticity to Certification by Analysis (CbA). However, populating an AeroDB requires evaluating a combinatorial space of geometries and operating conditions, making high-fidelity computational fluid dynamics (CFD) prohibitively expensive. The scale of this cost is evident in NASA's Space Launch System databases, which required thousands of CFD simulations and tens of millions of CPU-core hours \citep{nasa_sls_2017,nasa_sls_2018}, and the NASA X-57 aerodynamic database, which involved thousands of simulations with meshes containing hundreds of millions of cells \citep{nasa_x57_2021,frederick2025x57}. These computational demands have motivated neural surrogates that amortize simulation cost by learning the underlying input--output mapping \citep{brunton2020machine}. While recent surrogates can substantially accelerate CFD prediction, many downstream aerodynamic applications require only a small set of QoIs rather than complete high-dimensional flow fields, motivating reusable geometry representations that directly support physical prediction and downstream design tasks.

A large body of recent work has developed neural surrogates for predicting
high-dimensional physical fields on complex engineering geometries.
Transformer-, mesh- and geometry-aware architectures such as Transolver,
DoMINO, GeoTransolver and AB-UPT have demonstrated accurate prediction of
physical responses on general engineering domains
\citep{wu2024transolver,ranade2025domino,adams2025geotransolver,alkin2025abupt}.
These approaches are particularly valuable when the complete physical field
is required. However, aerodynamic design workflows also require repeated
evaluation of integrated quantities such as lift, drag and moments across
large design spaces. More fundamentally, a surrogate optimized for a
particular physical response does not necessarily provide a representation
that can be reused when operating conditions or the physical quantity of
interest changes. This motivates learning a reusable geometry representation that can support multiple downstream physical prediction and design tasks beyond a single geometry-to-field mapping.

Joint-Embedding Predictive Architectures (JEPAs) provide a natural framework
for learning such representations. Rather than reconstructing high-dimensional
observations directly, JEPA learns predictive relationships between latent
representations \citep{assran2023ijepa,bardes2024vjepa}. Related predictive
representation methods have likewise demonstrated that useful information
can be retained in compact latent spaces without explicit reconstruction
\citep{grill2020byol,baevski2022data2vec}. For engineering surrogate modeling,
however, the representation must remain useful as geometry, operating
conditions and physical response spaces change. A representation that
entangles these factors may be effective for a particular prediction task but
difficult to reuse when the operating condition or physical quantity of
interest changes. More broadly, self-supervised representation learning has
identified representation collapse as a potential failure mode, motivating
mechanisms such as redundancy reduction and variance preservation
\citep{zbontar2021barlow,bardes2022vicreg,caron2021dino,balestriero2025lejepa}.
Recent work has begun to explore predictive latent representations for
aerodynamic surrogate modeling. AeroJEPA, for example, predicts aerodynamic
latent representations from geometry and operating conditions and demonstrates
that such latent spaces can support linear probing and design-oriented
operations \citep{giral2026aerojepa}. However, its representation remains tied
to the aerodynamic prediction setting and operating conditions, leaving open a
more general question: can geometry itself be represented independently of
operating conditions and physical response, while retaining information that
is explicitly recoverable for downstream design?

We propose D-JEPA (Design-recoverable JEPA), a joint-embedding architecture
that separates geometry representation from physics prediction by encoding
each geometry into a compact latent space and conditioning lightweight physics
decoders on operating conditions. The geometry encoder receives neither
operating conditions nor target physics, while an explicit
design-recoverability objective encourages the latent to preserve geometric
design information. This factorization enables the learned representation to
be evaluated for predictive accuracy, design-variable recoverability and reuse
across operating conditions and physical response spaces. We demonstrate these
capabilities through full-field prediction, condition and cross-physics
transfer, as well as differentiable design optimization validated with
high-fidelity CFD. Our contributions are as follows:

\begin{enumerate}
\item A geometry-centered JEPA formulation for scientific surrogate
      modeling that explicitly factorizes a reusable geometry representation
      from operating-condition- and physics-specific prediction.
\item An explicit design-recoverability objective that encourages the
      shared geometry latent to preserve information about the underlying
      geometric design variables, making this information directly
      recoverable through a shallow probe.
\item A reusable geometry--physics factorization in which the geometry
      encoder is shared across operating conditions and physical response
      spaces, while lightweight physics-specific decoders can be swapped
      without retraining the geometry encoder.
\item Empirical evaluation of the resulting representation across
      parameterized aerodynamic benchmarks, operating-condition reuse,
      cross-physics transfer and differentiable design optimization with
      high-fidelity CFD validation.
\end{enumerate}

\section{Method}
In this section, we present the mathematical formulation of D-JEPA, including the geometry representation, latent prediction, physics-specific decoding and design-recoverability objective.

\subsection{Problem Formulation}

We consider the problem of predicting steady physical fields on a set of parameterized geometries under varying operating conditions and multiple sets of governing equations. Let a design be defined by its boundary surface $\partial\Omega$ embedded in $\Omega \subset \mathbb{R}^3$, represented in a canonical reference coordinate system. The surface is discretized as a point cloud $P = \{x_i\}_{i=1}^{N}$, with corresponding surface normals $N = \{n_i\}_{i=1}^{N}$. Let $c \in \mathbb{R}^{n_{\mathrm{cond}}}$ denote the operating environment, such as Reynolds number, angle of attack, Mach number or structural load cases. We index different physical regimes by $s \in \mathcal{S}$, where each regime corresponds to a particular set of governing equations, such as fluid or structural physics. The corresponding ground-truth field is denoted by $F_s:\Omega \rightarrow \mathbb{R}^{d_s}$.

Each geometry is associated with a design vector $\theta \in \mathbb{R}^{D}$. The design parameters are never provided as inputs to the physics decoder; instead, they are used during training only through an auxiliary design-recoverability objective. The primary objective is to learn a geometry encoder $E_{\vartheta}$ that maps the surface representation to a compact latent representation
\begin{equation}
g = E_{\vartheta}(P,N), \qquad g \in \mathbb{R}^{d_g},
\end{equation}
where $d_g=128$, together with lightweight physics-specific decoders $\{D_{\phi_s}\}$. Field prediction is given by
\begin{equation}
\hat{F}_s(q) = D_{\phi_s}(g,q,c),
\qquad
g = E_{\vartheta}(P,N),
\end{equation}
where $q$ denotes a query location. By structural construction, $g$ does not receive $c$ or $s$ as inputs. Thus, operating-condition and physics-specific information can enter the prediction only through the downstream decoder, while the geometry encoder is shared across conditions and physics regimes. This provides the factorization underlying D-JEPA: geometry is represented once in $g$, while operating conditions and physical response spaces are handled by the corresponding decoder.

\subsection{D-JEPA Framework}

\begin{figure}[t]
\centering
\includegraphics[width=12cm]{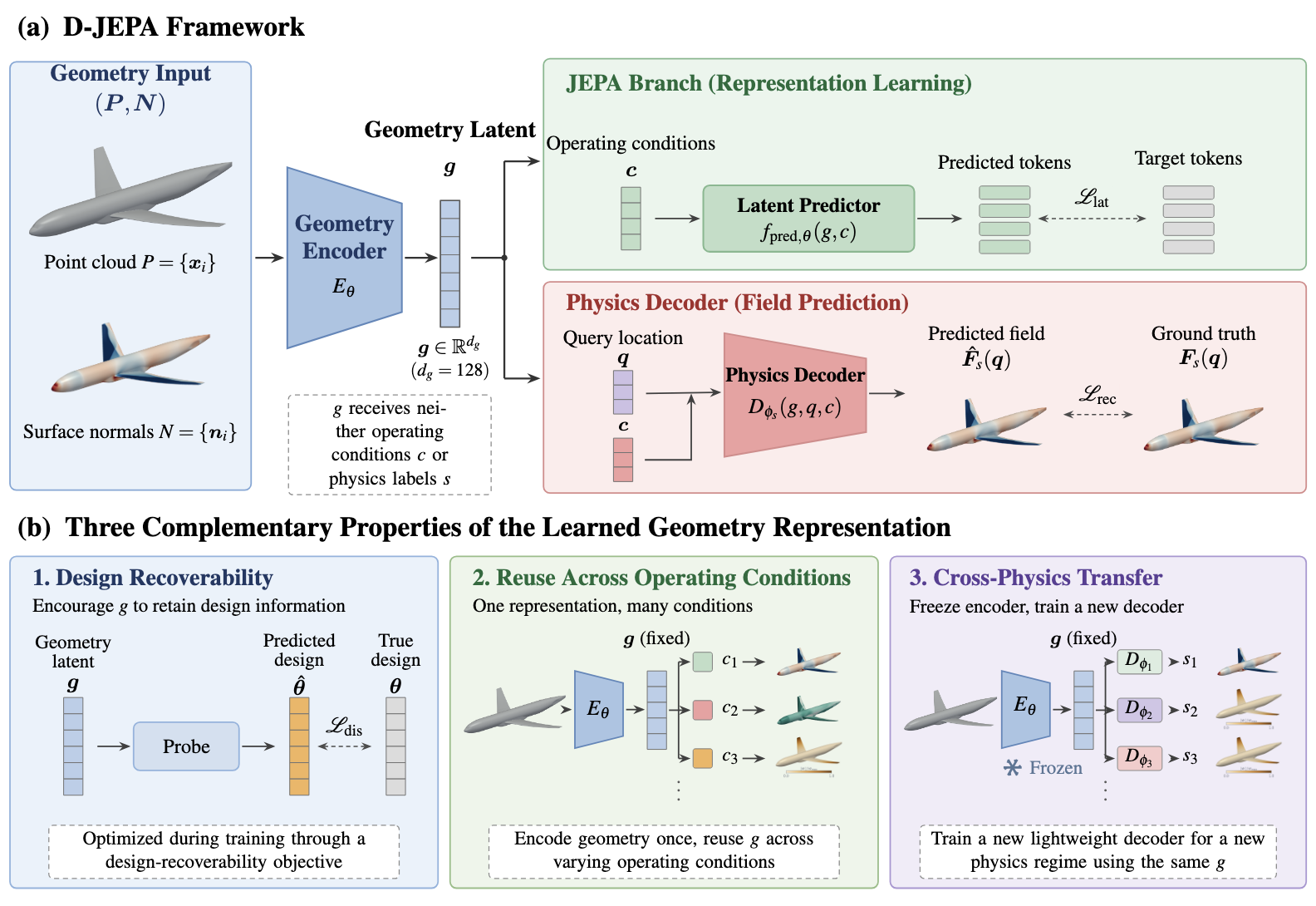}
\caption{Overview of D-JEPA: (a) D-JEPA architecture and training pathways. (b) Design recoverability, condition reuse and cross-physics transfer.}
\label{fig:overview}
\end{figure}

The D-JEPA framework separates geometry $g$, operating conditions $c$ and
governing physics through the decoder $\phi_s$, enabling design
recoverability, reuse across operating conditions and transfer across
physical response spaces (Figure~\ref{fig:overview}). During training, the
geometry encoder, target encoder, latent predictor, physics decoder and design
probe are jointly optimized. The target encoder $E_{\mathrm{target}}$ receives
an independently sampled ground-truth field point cloud and produces target
tokens $Z_t$, while the latent predictor
$f_{\mathrm{pred},\theta}(g,c)$ produces predicted tokens $\hat{Z}_t$.
Gradients from $\mathcal{L}_{\mathrm{lat}}$ propagate through both the
predictor and target encoder; no stop-gradient or EMA teacher is used.
The target encoder is used only during training and is discarded at
inference. The JEPA predictor provides an auxiliary latent prediction signal
that shapes the geometry representation, whereas the physics decoder
$D_{\phi_s}(g,q,c)$ is responsible for direct physical-field prediction. The two pathways therefore serve distinct roles. The JEPA pathway provides
representation-level supervision by requiring the geometry-dependent
prediction $\hat{Z}_t$ to match target tokens derived from the physical
field, while the physics decoder maps the shared geometry representation
directly to the physical field at arbitrary query locations. Thus,
$\mathcal{L}_{\mathrm{lat}}$ shapes the representation, whereas
$\mathcal{L}_{\mathrm{rec}}$ evaluates its utility for physical-field
prediction.

\subsection{Coordinate-Aware Geometry Encoding}

A local point-attention mechanism can capture geometric features such as local curvature but does not by itself provide an explicit representation of global proportions and orientation. We augment the learned point-transformer representation \citep{wu2024ptv3} with two parameter-free geometric descriptor blocks computed in the canonical reference frame. The first descriptor is a global descriptor $s_{\mathrm{glob}}$ with 15 dimensions. It contains the per-axis mean and standard deviation, bounding-box extent and the eigenvalues and off-diagonal entries of the $3\times3$ position covariance matrix. These features provide compact information about the global scale, extent and spatial distribution of the geometry. The second descriptor is a regional descriptor $s_{\mathrm{reg}}$ obtained from a $5\times5\times5$ voxel grid. Each voxel contains 13 statistics summarizing the mean and standard deviation of positions, normal statistics and occupancy, resulting in a $1625$-dimensional descriptor. In parallel, a learned point-transformer branch tokenizes the point cloud into $M=512$ tokens with width $d=128$, following the hierarchical point-set construction of \citet{qi2017pointnet2}. Mean and max pooling produce $s_{\mathrm{loc}} \in \mathbb{R}^{256}$. Layer sizes and neighbourhood parameters are given in Appendix~\ref{app:arch}. The local, global and regional descriptors are concatenated, $[s_{\mathrm{loc}};s_{\mathrm{glob}};s_{\mathrm{reg}}] \in \mathbb{R}^{1896}$, and passed through an MLP to obtain the geometry latent $g \in \mathbb{R}^{128}$. The resulting latent receives neither operating conditions $c$ nor physics labels $s$, and the same geometry representation is used by every downstream physics decoder and operating point.

\subsection{FiLM-Modulated Implicit Neural Field Decoder}

For each physics regime $s$, the decoder maps a query location $q$ to the corresponding local physical field and is modulated by the geometry latent $g$ through Feature-wise Linear Modulation (FiLM) \citep{perez2018film}. This follows the implicit neural representation formulation used for continuous fields \citep{sitzmann2020siren,mildenhall2021nerf}. The query location is encoded using a Fourier feature embedding with 16 octaves \citep{tancik2020fourier}, producing a $99$-dimensional representation. The decoder additionally receives the local surface normal $n(q)$ and operating condition $c$. The decoder consists of four residual FiLM blocks. For each block $l$, the latent representation generates the modulation parameters
\begin{equation}
(\gamma_l,\beta_l) = \mathrm{split}(A_l g),
\end{equation}
and the hidden representation is updated as
\begin{equation}
h_{l+1}
=
h_l
+
\mathrm{GELU}
\left(
\gamma_l \odot W_l\mathrm{LayerNorm}(h_l)
+
\beta_l
\right).
\end{equation}
The predicted field is obtained from the final hidden representation as
\begin{equation}
\hat{F}_s(q) = W_{\mathrm{out}}h_L.
\end{equation}

This formulation allows the same geometry representation to condition different physics-specific decoders while operating conditions enter explicitly through the decoder. Automatic differentiation provides surrogate gradients with respect to the latent and design variables, without requiring a separate PDE-adjoint computation.

\subsection{Training Objectives and Design Recoverability}

During Stage 1, the geometry encoder, latent predictor, primary physics decoder and design probe are jointly optimized. A target encoder $E_{\mathrm{target}}$ receives an independently sampled ground-truth field point cloud and produces target tokens
\begin{equation}
Z_t \in \mathbb{R}^{M\times d}.
\end{equation}
The predictor maps $(g,c)$ to predicted target tokens $\hat{Z}_t$. An auxiliary target decoder $D_t$ reconstructs the physical field from $Z_t$, while the primary physics decoder $D_{\phi_s}$ predicts the field from the geometry representation $g$. The composite training objective is
\begin{equation}
\mathcal{L}
=
\mathcal{L}_{\mathrm{lat}}
+
\lambda_r\mathcal{L}_{\mathrm{rec}}
+
\lambda_d\mathcal{L}_{\mathrm{dis}}
+
\lambda_s\mathcal{L}_{\mathrm{sig}}
+
\lambda_c\mathcal{L}_{\mathrm{sig\text{-}case}}
+
\lambda_v\mathcal{L}_{\mathrm{var}}
+
\lambda_t\mathcal{L}_{\mathrm{rec\text{-}t}}.
\end{equation}

The latent prediction loss is $\mathcal{L}_{\mathrm{lat}}
=
\left\|
\hat{Z}_t-Z_t
\right\|_2^2$ and the primary field reconstruction loss is $\mathcal{L}_{\mathrm{rec}}
=
\mathbb{E}_{q}
\left[
\left\|
D_{\phi_s}(g,q,c)-F_s(q)
\right\|_2^2
\right]$. To explicitly encourage the geometry representation to preserve information about the underlying design, we introduce a design-recoverability loss
\begin{equation}
\mathcal{L}_{\mathrm{dis}}
=
\left\|
\mathrm{probe}(g)-\theta
\right\|_2^2.
\end{equation}
The design parameters are never supplied to the physics decoder. Instead, they supervise the representation through $\mathcal{L}_{\mathrm{dis}}$, explicitly encouraging $g$ to retain design-relevant information. Unlike a purely post-hoc probing analysis \citep{alain2017probes}, design recoverability is therefore directly optimized during representation learning.

\subsection{Preventing Representation Collapse}

As design recoverability requires the representation to distinguish
different geometries, we additionally constrain variation across design cases.
Standard SIGReg \citep{balestriero2025lejepa} operates on the flattened token
ensemble and thus does not guarantee diversity across cases. In our setting,
representations can exhibit a near-constant token pattern across designs, with
between-case standard deviation of approximately $0.0032$, while satisfying
the token-level SIGReg constraint. We address this failure mode with three
complementary terms. First, case-level SIGReg is applied to the mean target
representation
\begin{equation}
\mathcal{L}_{\mathrm{sig\text{-}case}}
=
\mathrm{SIGReg}
\left(
\frac{1}{M}\sum_m Z_{t,m}
\right).
\end{equation}
Second, we explicitly constrain the variance of the representation across design cases:
\begin{equation}
\mathcal{L}_{\mathrm{var}}
=
\frac{1}{d}
\sum_j
\mathrm{ReLU}
\left(
\gamma
-
\sqrt{
\mathrm{var}_{\mathrm{case}}
(\bar{Z}_t)_j+\epsilon
}
\right).
\end{equation}
Finally, we require the target representation to retain sufficient information to reconstruct the physical field:
\begin{equation}
\mathcal{L}_{\mathrm{rec\text{-}t}}
=
\mathbb{E}_q
\left[
\left\|
D_t(Z_t,q)-F_s(q)
\right\|_2^2
\right].
\end{equation}

We set $\lambda_c=\lambda_v=\lambda_t=1$. Together, these constraints increase the observed between-case standard
deviation to approximately $1.0$ across the evaluated datasets, preventing the
representation from satisfying token-level diversity constraints while
remaining effectively constant across geometries.

\subsection{Physics Decoder Transfer}

After Stage 1, the geometry encoder is frozen and a new physics decoder $\phi_{s'}$ is trained for a new physical response space. This follows the parameter-efficient adaptation strategy used elsewhere in representation learning, where a frozen backbone is paired with a small trainable module \citep{houlsby2019adapters,hu2022lora}. Given $g=E_{\vartheta}(P,N)$, where gradients are not propagated through $E_{\vartheta}$, the transferred decoder is optimized according to
\begin{equation}
\min_{\phi_{s'}}
\mathbb{E}
\left[
\left\|
D_{\phi_{s'}}(g,q,c)-F_{s'}(q)
\right\|_2^2
\right].
\end{equation}

Because the geometry representation is fixed, this experiment tests whether the latent captures geometric information that can be reused by a new physical decoder. Transfer efficiency therefore depends on whether \(g\) encodes reusable geometric features rather than task-specific information.

\section{Results \& Discussion}

We evaluate the learned geometry representation along three properties: (1) predictive accuracy for full-field and integrated quantities, (2) design information retention and reuse across operating conditions, and (3) transfer to new physical response spaces through swappable decoders. Finally, we evaluate differentiable design optimization with high-fidelity CFD validation.

\subsection{Experimental datasets}

We evaluate D-JEPA on four complementary 3D benchmarks. SHIFT-Wing
\citep{shift_wing_2025} provides highly parameterized 3D aircraft wing
geometries spanning transonic flow regimes, with varied aspect ratios, sweep
and spanwise twist distributions. SHIFT-Submarine \citep{shift_submarine_2026}
contains parameterized axisymmetric and non-axisymmetric submarine hulls
featuring variations in fore-body shape, sail geometry and tail half-angles.
SHIFT-SUV \citep{shift_suv_2025} covers parameterized automotive geometries
that capture complex bluff-body aerodynamic phenomena, including slant angle,
backlight geometry and underbody tapering. Finally, AutoHood 3D
\citep{sharma2025autohood3d} pairs automotive hood geometries with coupled
fluid--structure interaction (FSI) data, in which aerodynamic surface pressure
fields act as mechanical loads that generate structural deformation responses.

\subsection{Full-field surrogate accuracy}

We establish that D-JEPA retains strong full-field-prediction accuracy
while introducing an explicitly structured geometry representation. Table~\ref{tab:main}
reports the coefficient of determination ($R^2$) for surface pressure, the
three components of wall shear stress (WSS$_x$, WSS$_y$, WSS$_z$) and the
integrated force coefficients ($C_D$, $C_L$), for D-JEPA and three baselines:
the AeroJEPA predictive-latent model \citep{giral2026aerojepa} and the
field-reconstruction surrogates GeoTransolver \citep{adams2025geotransolver}
and DoMINO \citep{ranade2025domino}. Implementation details for the baselines
are given in Appendix~\ref{app:baselines}. Across all three aerodynamic benchmarks, D-JEPA maintains or improves
field-prediction accuracy relative to AeroJEPA. Surface pressure prediction
reaches $R^2=0.9971$ on SHIFT-Wing, $0.9967$ on SHIFT-Submarine and
$0.9937$ on SHIFT-SUV. For wall shear stress, D-JEPA improves across every
component, with WSS$_y$ and WSS$_z$ improving on all benchmarks. Relative to
the full baseline set, D-JEPA attains the best value in all applicable
dataset--channel combinations. D-JEPA also preserves accuracy for integrated aerodynamic observables. On
SHIFT-Wing, $C_D$ prediction improves from $R^2=0.973$ with AeroJEPA to
$0.990$, while $C_L$ improves from $0.952$ to $0.970$. On SHIFT-SUV,
D-JEPA matches the best $C_D$ prediction at $R^2=1.000$. These results show
that the structured representation preserves the predictive performance
required for both spatially resolved fields and integrated engineering
quantities.

\begin{table}[t]
\centering
\caption{Surrogate accuracy across parameterized 3D aerodynamic benchmarks.
Full-field $R^2$ is reported for surface pressure and wall-shear-stress
components, together with integrated force coefficients $C_D$ and $C_L$.
Values are mean $\pm$ std over three seeds; bold indicates the best value for
each dataset and quantity.}
\label{tab:main}
\small
\setlength{\tabcolsep}{4pt}
\resizebox{\textwidth}{!}{%
\begin{tabular}{llcccccc}
\toprule
\textbf{Model} & \textbf{Dataset} & \textbf{Pressure} & \textbf{WSS$_x$} &
\textbf{WSS$_y$} & \textbf{WSS$_z$} & \textbf{$C_D$} & \textbf{$C_L$} \\
\midrule
\multirow{3}{*}{DoMINO}
 & SHIFT-Wing      & 0.9678 $\pm$ 0.001 & 0.8972 $\pm$ 0.001 & 0.9456 $\pm$ 0.004 & 0.9654 $\pm$ 0.002 & 0.940 $\pm$ 0.002 & 0.891 $\pm$ 0.001 \\
 & SHIFT-Submarine & 0.9321 $\pm$ 0.000 & 0.9332 $\pm$ 0.000 & 0.9421 $\pm$ 0.003 & 0.9243 $\pm$ 0.002 & 0.973 $\pm$ 0.001 & -- \\
 & SHIFT-SUV       & 0.9765 $\pm$ 0.000 & 0.9475 $\pm$ 0.000 & 0.9241 $\pm$ 0.002 & 0.9342 $\pm$ 0.002 & 0.980 $\pm$ 0.003 & -- \\
\midrule
\multirow{3}{*}{GeoTransolver}
 & SHIFT-Wing      & 0.9950 $\pm$ 0.001 & 0.9350 $\pm$ 0.001 & 0.9645 $\pm$ 0.002 & 0.9899 $\pm$ 0.001 & 0.960 $\pm$ 0.002 & 0.956 $\pm$ 0.001 \\
 & SHIFT-Submarine & 0.9921 $\pm$ 0.001 & 0.9432 $\pm$ 0.001 & 0.9621 $\pm$ 0.001 & 0.9743 $\pm$ 0.002 & 0.990 $\pm$ 0.002 & -- \\
 & SHIFT-SUV       & 0.9931 $\pm$ 0.001 & 0.9698 $\pm$ 0.001 & 0.9641 $\pm$ 0.001 & 0.9542 $\pm$ 0.001 & 0.996 $\pm$ 0.002 & -- \\
\midrule
\multirow{3}{*}{AeroJEPA}
 & SHIFT-Wing      & 0.9892 $\pm$ 0.009 & 0.9168 $\pm$ 0.003 & 0.9618 $\pm$ 0.007 & 0.9719 $\pm$ 0.007 & 0.973 $\pm$ 0.011 & 0.952 $\pm$ 0.046 \\
 & SHIFT-Submarine & 0.9937 $\pm$ 0.000 & 0.9532 $\pm$ 0.001 & 0.9602 $\pm$ 0.002 & 0.9764 $\pm$ 0.001 & 0.987 $\pm$ 0.003 & -- \\
 & SHIFT-SUV       & 0.9870 $\pm$ 0.001 & 0.9352 $\pm$ 0.001 & 0.9522 $\pm$ 0.001 & 0.9437 $\pm$ 0.001 & \ 1.000 $\pm$ 0.000 & -- \\
\midrule
\multirow{3}{*}{\textbf{D-JEPA}}
 & SHIFT-Wing      & \textbf{0.9971 $\pm$ 0.001} & \textbf{0.9376 $\pm$ 0.002} & \textbf{0.9849 $\pm$ 0.001} & \textbf{0.9900 $\pm$ 0.002} & \textbf{0.990 $\pm$ 0.002} & \textbf{0.970 $\pm$ 0.005} \\
 & SHIFT-Submarine & \textbf{0.9967 $\pm$ 0.001} & \textbf{0.9691 $\pm$ 0.000} & \textbf{0.9766 $\pm$ 0.000} & \textbf{0.9809 $\pm$ 0.001} & \textbf{0.995 $\pm$ 0.002} & -- \\
 & SHIFT-SUV       & \textbf{0.9937 $\pm$ 0.001} & \textbf{0.9747 $\pm$ 0.002} & \textbf{0.9649 $\pm$ 0.003} & \textbf{0.9782 $\pm$ 0.002} & \textbf{1.000 $\pm$ 0.000} & -- \\
\bottomrule
\end{tabular}}
\end{table}

\subsection{Geometry representation and transfer capabilities}

Having established full-field surrogate accuracy, we evaluate the properties that distinguish D-JEPA from a conventional predictive surrogate: case-level representation diversity, design-variable recoverability, reuse across operating conditions and cross-physics decoder transfer.

\subsubsection{Preventing case-level representation collapse}

Design recoverability requires the latent to distinguish different geometries,
but token-level diversity alone does not guarantee variation across design
cases. We therefore evaluate the effect of the representation constraints on
between-case variation. Table~\ref{tab:collapse} summarizes this ablation.
Without case-level constraints, the between-case standard deviation of the
target representation remains near zero ($0.0002$--$0.0004$) across all three
benchmarks. Adding token-level SIGReg alone produces essentially no change,
showing that token-level diversity does not prevent case-level collapse.
Case-level constraints substantially increase between-case variation. The
variance hinge and target-reconstruction terms increase the standard deviation
to approximately $1.0$ across all three geometry families, reaching
$1.03$--$1.06$ on SHIFT-Wing and SHIFT-Submarine and approximately $0.99$ on
SHIFT-SUV. Importantly, these changes occur with only modest changes in
forward pressure prediction, indicating that the additional constraints
primarily alter the structure of the representation rather than its predictive
capacity. The complete diagnostics, including latent and reconstruction
losses, are reported in Appendix~\ref{app:collapse}.

\begin{table}[t]
\centering
\caption{Case-level variation in the target representation under progressively
stronger representation constraints. Values report
$\mathrm{std}_{\mathrm{case}}(\mathrm{mean}_m Z_t)$, measuring variation
across geometry cases.}
\label{tab:collapse}
\scriptsize
\setlength{\tabcolsep}{4pt}
\begin{tabular}{lccccc}
\toprule
\textbf{Dataset}
& \textbf{None}
& \textbf{Token SIGReg}
& \textbf{Case SIGReg}
& \textbf{+ Variance}
& \textbf{+ Target Recon} \\
\midrule
SHIFT-Wing
& 0.0003 & 0.0003 & 0.6211 & 1.0346 & 1.0317 \\
SHIFT-Submarine
& 0.0004 & 0.0005 & 0.7566 & 1.0633 & 1.0620 \\
SHIFT-SUV
& 0.0002 & 0.0002 & 0.0003 & 0.9892 & 0.9889 \\
\bottomrule
\end{tabular}
\end{table}

\subsubsection{Design-variable recoverability}

A central objective of D-JEPA is to retain information about the underlying
geometric design variables in the learned latent. We therefore evaluate
whether the design parameters can be recovered using a linear probe from
$g$. Table~\ref{tab:recover} in Appendix~\ref{app:supp} reports recovery
$R^2$ for every design variable of SHIFT-SUV, SHIFT-Wing and
SHIFT-Submarine. D-JEPA achieves near-perfect recoverability across the parameterized shape
dimensions, with mean $R^2$ scores of 0.989 on SHIFT-SUV, 0.998 on SHIFT-Wing
and 0.998 on SHIFT-Submarine. As a representation-level reference, we apply
the identical probe to the AeroJEPA context latent, mean-pooled to the same
128 dimensions as $g$. The corresponding mean $R^2$ values are 0.683,
0.801 and 0.882, respectively, giving gains of 0.306, 0.197 and 0.116.
Thus, D-JEPA retains substantially more linearly recoverable design
information across all three geometry families. This structured decodability provides a direct representation-level basis for
design navigation and complements the full-field prediction results: the latent
is not only useful for predicting physical fields, but also retains explicit
information about the geometric variables defining the input shapes. We further perform component ablations to isolate the sources of this
recoverability. Here \emph{Learned-only} denotes the point-transformer branch
alone, with the parameter-free global and regional descriptors removed, so
that the comparison isolates the contribution of the coordinate-aware
descriptors. As shown in Table~\ref{tab:ablation}, removing the design
loss $\mathcal{L}_{\mathrm{dis}}$ substantially reduces design recovery across
all three datasets, while having a comparatively smaller effect on pressure
prediction. Removing the latent prediction loss $\mathcal{L}_{\mathrm{lat}}$
produces a smaller but consistent reduction in design and pressure $R^2$
across the three datasets. The learned-only encoder retains substantial design
information, indicating that recoverability is not solely due to the
explicit geometric descriptors.

\begin{table}[t]
\centering
\caption{Component ablation of D-JEPA across the SHIFT datasets, showing the contributions of latent prediction and design supervision to design recoverability and pressure prediction. \emph{Learned-only} removes the parameter free global and regional descriptors.}
\label{tab:ablation}
\scriptsize
\setlength{\tabcolsep}{3pt}
\begin{tabular}{lcccc|cccc}
\toprule
& \multicolumn{4}{c|}{\textbf{Design R$^2$} $\uparrow$}
& \multicolumn{4}{c}{\textbf{Pressure R$^2$} $\uparrow$} \\
\textbf{Dataset}
& \textbf{Full}
& \textbf{$-\mathcal{L}_{\mathrm{lat}}$}
& \textbf{Learned-only}
& \textbf{$-\mathcal{L}_{\mathrm{dis}}$}
& \textbf{Full}
& \textbf{$-\mathcal{L}_{\mathrm{lat}}$}
& \textbf{Learned-only}
& \textbf{$-\mathcal{L}_{\mathrm{dis}}$} \\
\midrule
SHIFT-Wing
& \textbf{0.998} & 0.993 & 0.976 & 0.722
& \textbf{0.9971} & 0.9912 & 0.9885 & 0.9823 \\
SHIFT-Submarine
& \textbf{0.998} & 0.982 & 0.757 & 0.564
& \textbf{0.9967} & 0.9896 & 0.9852 & 0.9872 \\
SHIFT-SUV
& \textbf{0.989} & 0.984 & 0.976 & 0.622
& \textbf{0.9937} & 0.9851 & 0.9801 & 0.9821 \\
\bottomrule
\end{tabular}
\end{table}

\subsubsection{Operating-condition sweeps (same physics, new conditions)}

As the geometry encoder $E_\vartheta$ receives neither the condition vector
$c$ nor the physics label $s$, the latent representation $g$ can be computed
once per geometry and reused across operating conditions. We evaluate this
capability on SHIFT-Wing across a transonic Mach sweep
($\mathrm{Mach}\in[0.50,0.90]$). Ten Mach values are used during training,
while Mach $=0.7750$ and Mach $=0.8125$ are held out entirely. Importantly,
both held-out values lie within the trained Mach range and therefore test
\emph{interpolation} across operating conditions rather than extrapolation.
Table~\ref{tab:mach} reports pressure prediction $R^2$ across all
trained and held-out conditions. The frozen geometry representation maintains
$R^2\geq0.9902$ across all evaluated conditions, including $R^2=0.9922$ and
$0.9944$ at the two held-out Mach numbers. A single geometry representation
therefore remains reusable across the continuous operating sweep without
additional geometry-encoder training.

\begin{table}[t]
\centering
\caption{SHIFT-Wing pressure prediction accuracy ($R^2$) across trained and
held-out Mach numbers using a single frozen geometry representation $g$.
The held-out Mach values are interpolation points within the trained Mach
range. Bold values denote held-out conditions.}
\label{tab:mach}
\small
\setlength{\tabcolsep}{3.5pt}
\resizebox{\textwidth}{!}{%
\begin{tabular}{l cccccccccc cc}
\toprule
& \multicolumn{10}{c}{\textbf{Trained operating conditions}}
& \multicolumn{2}{c}{\textbf{Held-out}} \\
\cmidrule(lr){2-11}\cmidrule(lr){12-13}
\textbf{Mach}
& 0.5000 & 0.6000 & 0.7000 & 0.7500 & 0.8000 & 0.8250 & 0.8375 & 0.8500
& 0.8750 & 0.9000 & \textbf{0.7750} & \textbf{0.8125} \\
\midrule
\textbf{Pressure $R^2$}
& 0.9902 & 0.9935 & 0.9948 & 0.9944 & 0.9941 & 0.9956 & 0.9960 & 0.9967
& 0.9952 & 0.9939 & \textbf{0.9922} & \textbf{0.9944} \\
\textbf{Std}
& 0.001 & 0.002 & 0.001 & 0.003 & 0.002 & 0.001 & 0.001 & 0.002
& 0.001 & 0.002 & 0.001 & 0.002 \\
\bottomrule
\end{tabular}}
\end{table}

\subsubsection{Swappable physics decoders (cross-physics transfer)}

We next test whether the same geometry representation can be reused when the
target physical response changes. We evaluate this capability on AutoHood 3D
by transferring a geometry encoder pretrained on aerodynamic pressure fields
to structural displacement prediction. In the D-JEPA transfer setting, the pretrained aerodynamic geometry encoder is
completely frozen and only a lightweight structural FiLM decoder is trained,
requiring 0.56\,M trainable parameters. In contrast, the from-scratch baseline
trains the full geometry encoder and structural decoder end-to-end, requiring
2.30\,M trainable parameters. Table~\ref{tab:crossphysics} tracks structural
prediction $R^2$ across training budgets. The swappable-decoder formulation reduces trainable parameters by $4.1\times$
and provides substantially faster adaptation at low training budgets. After
10 epochs, the transferred representation reaches $R^2=0.3316$, compared with
$0.0396$ from scratch. At 100 epochs, it reaches $0.8901$ versus $0.8740$,
and the transferred model remains ahead through 300 epochs, by which point the
two approaches are within $0.007$ of each other. The benefit demonstrated here
is therefore parameter-efficient and rapid adaptation: the aerodynamic geometry
representation can be reused for this CFD-to-structural transfer while
training only a lightweight physics-specific decoder.

\begin{table}[t]
\centering
\caption{Structural displacement prediction accuracy ($R^2$) on AutoHood 3D
comparing frozen geometry encoder transfer versus training from scratch.}
\label{tab:crossphysics}
\small
\resizebox{\textwidth}{!}{%
\begin{tabular}{lcc}
\toprule
\textbf{Training epoch budget} &
\textbf{Frozen geometry encoder (0.56\,M params)} &
\textbf{From scratch (2.30\,M params)} \\
\midrule
10  & \textbf{0.3316 $\pm$ 0.020} & 0.0396 $\pm$ 0.007 \\
100 & \textbf{0.8901 $\pm$ 0.013} & 0.8740 $\pm$ 0.034 \\
200 & \textbf{0.9327 $\pm$ 0.002} & 0.9297 $\pm$ 0.012 \\
300 & \textbf{0.9385 $\pm$ 0.001} & 0.9313 $\pm$ 0.002 \\
\bottomrule
\end{tabular}}
\end{table}

\subsection{Design optimization \& CFD validation}

Finally, we evaluate the representation in a downstream engineering workflow using gradient-based design optimization, with selected designs validated against high-fidelity CFD. The differentiable optimization path and design-space constraints are detailed in Appendix~\ref{app:opt}. To assess whether surrogate-guided optimization produces physically consistent
design rankings, we perform direct high-fidelity CFD validation for
SHIFT-Submarine and SHIFT-Wing. Equivalent high-fidelity CFD validation for
SHIFT-SUV was not possible because the corresponding high-fidelity CFD setup
and sufficient geometry and operating-condition information were not available.
For each benchmark, we evaluate four candidates selected from the optimization
trajectories. \textit{OPT-$\tau$} is the predicted best design within the
nominal design radius $\tau$, representing a ``safe'' optimum. \textit{OPT-Wide}
is the predicted best design with the radius relaxed to $1.8\tau$, explicitly
testing extrapolation beyond the nominal region. \textit{ANTI} is the predicted
worst design within $\tau$ and serves as a negative control, while \textit{MID}
is a held-out design with a mid-range predicted objective and provides a
buildable sanity check. For SHIFT-Submarine, CFD is evaluated at $\rho=998.2\,\mathrm{kg/m^3}$,
$V=5.0\,\mathrm{m/s}$, $q=12{,}478\,\mathrm{Pa}$ and
$A_{\mathrm{ref}}=19.27\,\mathrm{m^2}$. Predicted drag agrees with CFD within
$0.6$--$2.0\%$. For SHIFT-Wing, the relative error in integrated lift ranges
from $0.72$--$10.74\%$. Despite these differences in absolute error, the
surrogate exactly preserves the CFD ranking of all four candidates in both
benchmarks, with rankings $1\rightarrow1$, $2\rightarrow2$, $3\rightarrow3$
and $4\rightarrow4$, as reported in Table~\ref{tab:cfd}. These results provide
a direct high-fidelity check that the surrogate preserves the relative ordering
of optimized and control designs, including the extrapolative
\textit{OPT-Wide} case.

\begin{table}[h]
\centering
\caption{High-fidelity CFD validation of surrogate-optimized designs.
Integrated drag is reported for SHIFT-Submarine and integrated lift for
SHIFT-Wing. Predicted and CFD rankings agree for all evaluated designs.}
\label{tab:cfd}
\small
\setlength{\tabcolsep}{5pt}
\resizebox{\textwidth}{!}{%
\begin{tabular}{llcccc}
\toprule
\textbf{Benchmark} & \textbf{Design candidate} & \textbf{Predicted [N]} &
\textbf{CFD [N]} & \textbf{Relative error} &
\textbf{Predicted $\to$ CFD rank} \\
\midrule
\multirow{4}{*}{\begin{tabular}[c]{@{}l@{}}SHIFT-Submarine\\ (drag)\end{tabular}}
 & OPT-wide & 645.8 & 637.69 & 1.3\% & \textbf{1 $\to$ 1} \\
 & OPT-tau  & 677.0 & 688.36 & 1.7\% & \textbf{2 $\to$ 2} \\
 & MID      & 718.6 & 723.08 & 0.6\% & \textbf{3 $\to$ 3} \\
 & ANTI     & 758.7 & 773.75 & 2.0\% & \textbf{4 $\to$ 4} \\
\midrule
\multirow{4}{*}{\begin{tabular}[c]{@{}l@{}}SHIFT-Wing\\ (lift)\end{tabular}}
 & OPT-Wide & 3{,}426{,}080 & 3{,}193{,}866 & 7.27\%  & \textbf{1 $\to$ 1} \\
 & OPT-Tau  & 3{,}108{,}536 & 3{,}086{,}237 & 0.72\%  & \textbf{2 $\to$ 2} \\
 & MID      & 2{,}008{,}248 & 2{,}249{,}932 & 10.74\% & \textbf{3 $\to$ 3} \\
 & ANTI     & 1{,}434{,}578 & 1{,}386{,}571 & 3.46\%  & \textbf{4 $\to$ 4} \\
\bottomrule
\end{tabular}}
\end{table}

The experiments show that D-JEPA provides a reusable geometry representation
for accurate simulation, transfer and design optimization, improving
full-field prediction over the evaluated field-reconstruction and
predictive-latent baselines while maintaining strong integral accuracy and
enabling reuse across design spaces, operating conditions and downstream
physics without retraining the geometry encoder. Design-variable recovery
further provides a differentiable path from the physical objective back to the
design parameters, which supports gradient-based design exploration. However,
the evaluation remains limited to a small set of aerodynamic and structural
benchmarks, with physics transfer demonstrated only in the CFD-to-structural
setting and optimization restricted to differentiable surrogates with
predefined objectives; broader multi-physics validation, diverse geometries
and constrained real-world design problems remain important directions for
future work.

\section{Conclusions}

We present D-JEPA, a geometry-centric representation learning framework
designed to separate reusable geometric information from physics- and
condition-specific prediction. Across three challenging parameterized
three-dimensional aerodynamic benchmarks, D-JEPA maintains or improves
full-field surrogate accuracy relative to the evaluated baselines
while maintaining high accuracy for integral aerodynamic quantities.
We show that the learned geometric representation can be reused across
different geometry families and operating conditions, and transferred to
a structural response task on AutoHood 3D without retraining the geometry
encoder. Finally, the learned representation supports gradient-based design
optimization, with the resulting candidates validated against high-fidelity
CFD. Together,
these results demonstrate that explicitly separating geometry representation
from physics-specific prediction can provide a reusable geometric
representation by design, enabling a single learned representation to
support multiple prediction tasks and downstream optimization.

\bibliographystyle{iclr2027_conference}
\bibliography{references}

\newpage
\appendix

\setcounter{table}{0}
\renewcommand{\thetable}{A.\arabic{table}}
\setcounter{figure}{0}
\renewcommand{\thefigure}{A.\arabic{figure}}
\setcounter{equation}{0}
\renewcommand{\theequation}{A.\arabic{equation}}

\section*{Appendix}
\addcontentsline{toc}{section}{Appendix}

This appendix provides additional details on dataset construction and
preprocessing, model architecture, training objectives, representation
collapse diagnostics, evaluation protocols, baseline implementations,
design-variable recovery, operating-condition generalization, cross-physics
transfer, differentiable optimization and computational requirements. It is
intended to provide sufficient implementation and evaluation detail to
reproduce the experiments reported in the main paper.

\section{Dataset Specifications and Preprocessing Pipelines}

We evaluate D-JEPA across four benchmark datasets spanning parameterized
three-dimensional physical systems: SHIFT-Wing
\citep{shift_wing_2025}, SHIFT-Submarine
\citep{shift_submarine_2026}, SHIFT-SUV
\citep{shift_suv_2025} and AutoHood 3D
\citep{sharma2025autohood3d}. Each case is represented by an unstructured
spatial point cloud $P$ with associated outward surface normals $N$,
solver-generated physical fields, underlying geometric design parameters
and external operating conditions when applicable. Table~\ref{tab:a1}
summarizes the benchmark characteristics.

\begin{table}[h]
\centering
\caption{Benchmark dataset properties, parameter dimensions, operating
conditions and surface targets. A dash indicates that the quantity is not
applicable.}
\label{tab:a1}
\small
\resizebox{\textwidth}{!}{%
\begin{tabular}{lcccl}
\toprule
\textbf{Dataset} &
\textbf{Cases} &
\textbf{Design parameters} &
\textbf{Operating conditions} &
\textbf{Surface fields} \\
\midrule
SHIFT-Wing
& 1{,}966
& 7
& angle of attack, Mach
& pressure, WSS$_x$, WSS$_y$, WSS$_z$ \\

SHIFT-Submarine
& 4{,}000
& 7
& fixed
& pressure, WSS$_x$, WSS$_y$, WSS$_z$ \\

SHIFT-SUV
& 1{,}996
& 7
& fixed
& pressure, WSS$_x$, WSS$_y$, WSS$_z$ \\

AutoHood 3D
& 12{,}000
& --
& load case
& pressure, displacement \\
\bottomrule
\end{tabular}}
\end{table}

SHIFT-Wing is parameterized by seven geometric variables comprising aspect
ratio, quarter-chord sweep angle $\Lambda_{c/4}$ and five spanwise twist
variables. SHIFT-Submarine contains seven parameters controlling
cross-sectional geometry, fore-body shape and length, mid-body length,
aft-tail half-angle, sail-length ratio and strake-length ratio.
SHIFT-SUV contains seven parameters describing hood height, hood-base
height, windshield inclination, total vehicle height, backlight angle,
rear tapering ratio and front-rear planform curvature.

The explicit design parameters are not provided to the physics decoder.
Instead, they are used only by the auxiliary design-recoverability
objective during training. At inference, the decoder receives the learned
geometry representation $g$, query location $q$, local surface normal
$n(q)$ and operating condition vector $c$. This separation allows the
design variables to be evaluated as information retained by the geometry
representation rather than as direct decoder inputs.

AutoHood 3D provides a cross-physics evaluation in which the same geometric
representation can support different physical response spaces. The
geometry is associated with an aerodynamic pressure field and a structural
displacement response. D-JEPA therefore uses the same geometry
representation while allowing physics-specific decoders to predict the
corresponding outputs. For SHIFT-Submarine and SHIFT-SUV, the operating
condition vector is empty because the evaluated flow conditions are fixed.
Operating-condition conditioning is explicitly evaluated on SHIFT-Wing
through variation in Mach number and angle of attack.

\section{Detailed Model Architecture}
\label{app:arch}

\subsection{Geometry encoder}

The geometry encoder $E_{\vartheta}$ maps an unstructured point cloud and
its surface normals to a compact geometry representation, as given in
Eq.~\eqref{eq:a1}:
\begin{equation}
g = E_{\vartheta}(P,N) \in \mathbb{R}^{128}.
\label{eq:a1}
\end{equation}

The encoder first processes the input geometry into $M=512$ context tokens
with feature dimension $d=128$. Local spatial neighborhoods use
$k_{\mathrm{aggregate}}=16$ nearest neighbors for message passing and
$k_{\mathrm{attn}}=16$ neighbors for local self-attention. The backbone
contains six transformer blocks with eight attention heads and an MLP
expansion factor of four. The resulting token matrix
$Z\in\mathbb{R}^{512\times128}$ is summarized using channel-wise mean and
maximum pooling, as given in Eq.~\eqref{eq:a2}:
\begin{equation}
s_{\mathrm{loc}}
=
\left[
\meanop_m Z_m \,;\,
\max_m Z_m
\right]
\in\mathbb{R}^{256}.
\label{eq:a2}
\end{equation}

The local representation is augmented with non-parametric global and
regional descriptors to provide explicit information about large-scale
geometry.

\subsection{Global and regional geometric descriptors}

The global descriptor $s_{\mathrm{glob}}$ contains 15 geometric statistics
computed directly from the spatial coordinates. These include per-axis
coordinate means and standard deviations, oriented bounding-box
dimensions, eigenvalues of the spatial covariance matrix and off-diagonal
covariance terms.

The regional descriptor $s_{\mathrm{reg}}$ discretizes the canonical
bounding domain into a $5 \times 5 \times 5$ voxel grid. For each voxel, local
coordinate statistics, surface-normal statistics and point-density
information are computed. The resulting descriptor contains
$125\times13=1{,}625$ features.

The three representations are concatenated as in Eq.~\eqref{eq:a3}:
\begin{equation}
s_{\mathrm{fused}}
=
[s_{\mathrm{loc}}\,;\,s_{\mathrm{glob}}\,;\,s_{\mathrm{reg}}]
\in\mathbb{R}^{1896},
\label{eq:a3}
\end{equation}
and passed through an MLP to produce the final 128-dimensional geometry
representation $g$.

Neither the operating condition vector $c$ nor the physics identifier $s$
is provided to the geometry encoder. Thus, the geometry representation is
computed solely from geometric information and can subsequently be reused
when the operating condition or physical response space changes.

\subsection{JEPA representation-learning pathway}

D-JEPA uses latent prediction as an auxiliary representation-learning
objective. During training, a context representation is used to predict
target latent tokens through the JEPA predictor. The resulting latent
prediction objective encourages the geometry representation to retain
information useful for predicting the target representation. The JEPA predictor is not used as the physics-field decoder. Instead, the
learned geometry representation $g$ is passed directly to the
physics-specific decoder for downstream field prediction. This separation
allows the JEPA pathway to shape the geometry representation while keeping
the physical output mapping independently swappable.

\subsection{FiLM-modulated physics decoder}

For a physical response space $s$, the decoder is written as in
Eq.~\eqref{eq:a4}:
\begin{equation}
\hat{F}_s(q)
=
D_{\phi_s}(g,q,c),
\label{eq:a4}
\end{equation}
where $g$ is the geometry representation, $q$ is the query location,
$n(q)$ is the local surface normal and $c$ denotes the operating
condition. Query coordinates are encoded using Fourier features across 16 frequency
octaves \citep{tancik2020fourier}, producing a 99-dimensional positional
representation. The decoder
contains four residual FiLM blocks with hidden width 256. The geometry
representation modulates the query pathway through learned affine scale
and shift parameters rather than being treated only as a concatenated
input \citep{perez2018film}. This factorization separates geometry
representation learning from the
mapping from geometry and operating conditions to a particular physical
response space. Consequently, a new decoder can be trained while keeping
the geometry encoder fixed. For the aerodynamic benchmarks, the decoder predicts four surface
quantities: pressure and the three components of wall shear stress
(WSS$_x$, WSS$_y$ and WSS$_z$).

\section{Training Objectives and Optimization}

\subsection{Composite training objective}

D-JEPA is trained end-to-end using a composite objective that combines
latent prediction, physical reconstruction, design-variable recovery,
representation regularization and target reconstruction, as given in
Eq.~\eqref{eq:a5}:
\begin{equation}
\mathcal{L} =
\mathcal{L}_{\mathrm{lat}}
+
\lambda_r\mathcal{L}_{\mathrm{rec}}
+
\lambda_d\mathcal{L}_{\mathrm{dis}}
+
\lambda_s\mathcal{L}_{\mathrm{sig}}
+
\lambda_c\mathcal{L}_{\mathrm{sig\text{-}case}}
+
\lambda_v\mathcal{L}_{\mathrm{var}}
+
\lambda_t\mathcal{L}_{\mathrm{rec\text{-}t}}.
\label{eq:a5}
\end{equation}

The latent prediction term $\mathcal{L}_{\mathrm{lat}}$ aligns predicted latent
representations with target-encoder representations. The physical
reconstruction loss $\mathcal{L}_{\mathrm{rec}}$ trains the physics decoder to
predict the target physical field. The design-recoverability loss
$\mathcal{L}_{\mathrm{dis}}$ trains a linear probe to recover the known design
variables from the geometry representation. The token-level SIGReg term
$\mathcal{L}_{\mathrm{sig}}$ \citep{balestriero2025lejepa} encourages diversity
within the target-token representation. Because token-level diversity does
not by itself guarantee variation between different geometry cases, D-JEPA
also evaluates case-level regularization. The case-level SIGReg term
$\mathcal{L}_{\mathrm{sig\text{-}case}}$ and variance constraint
$\mathcal{L}_{\mathrm{var}}$
encourage variation across geometry cases, while the target-reconstruction
term $\mathcal{L}_{\mathrm{rec\text{-}t}}$ provides an additional connection
between target representations and local physical coordinates. The resulting
objective is designed to jointly maintain predictive accuracy, design
information and non-collapsed geometry representations. The corresponding
weights are listed in Table~\ref{tab:a2}.

\begin{table}[h]
\centering
\caption{Hyperparameter weights used for the composite D-JEPA objective.}
\label{tab:a2}
\small
\setlength{\tabcolsep}{4pt}
\resizebox{\textwidth}{!}{%
\begin{tabular}{lccccccc}
\toprule
\textbf{Loss term}
&
Latent
&
Recon.
&
Design
&
Token
&
Case
&
Var.
&
Tgt.\ rec. \\
&
($\mathcal{L}_{\mathrm{lat}}$)
&
($\lambda_r\mathcal{L}_{\mathrm{rec}}$)
&
($\lambda_d\mathcal{L}_{\mathrm{dis}}$)
&
($\lambda_s\mathcal{L}_{\mathrm{sig}}$)
&
($\lambda_c\mathcal{L}_{\mathrm{sig\text{-}case}}$)
&
($\lambda_v\mathcal{L}_{\mathrm{var}}$)
&
($\lambda_t\mathcal{L}_{\mathrm{rec\text{-}t}}$)
\\
\midrule
\textbf{Weight}
&
1.0
&
1.0
&
1.0
&
0.01
&
1.0
&
1.0
&
1.0
\\
\bottomrule
\end{tabular}}
\end{table}

\subsection{Optimization schedule}

AeroJEPA and D-JEPA use matched optimization settings unless otherwise
specified. The primary training configuration is summarized in
Table~\ref{tab:a3}. The batch size is 16 for SHIFT-Wing, SHIFT-SUV and
SHIFT-Submarine, selected to accommodate the memory requirements of the
underlying solver meshes.

\begin{table}[h]
\centering
\caption{Optimization configuration and training hyperparameters.}
\label{tab:a3}
\small
\begin{tabular}{ll}
\toprule
\textbf{Setting} & \textbf{Value} \\
\midrule
Optimizer    & AdamW \citep{loshchilov2019adamw} \\
Base LR      & $1\times10^{-3}$ \\
Weight decay & $1\times10^{-3}$ \\
LR schedule  & Cosine decay \\
Gradient clipping & 10.0 \\
Epochs       & 400 \\
Precision    & FP32 \\
Seeds        & 42, 43, 44 \\
\bottomrule
\end{tabular}
\end{table}

\section{Representation-Collapse Diagnostics}
\label{app:collapse}

A central requirement of the geometry representation is that different
geometry cases should provide sufficiently distinct learning signals to
avoid case-level collapse. We therefore measure between-case variation in
the target representation using Eq.~\eqref{eq:a6}:
\begin{equation}
\sigma_{\mathrm{case}}
=
\stdop_{\mathrm{case}}
\left(
\meanop_m Z_t
\right),
\label{eq:a6}
\end{equation}
where $Z_t$ denotes the target token representation and the mean is taken
over tokens before computing the standard deviation across geometry cases.
This diagnostic distinguishes case-level variation from token-level
diversity: a target representation may contain diverse tokens within an
individual geometry while remaining nearly identical across different
geometries.

Table~\ref{tab:collapse_full} extends the summary in
Table~\ref{tab:collapse} with the latent and reconstruction losses and the
corresponding downstream metrics. Two observations are relevant beyond the
between-case standard deviation itself. First, the latent loss
$L_{\mathrm{lat}}$ is smallest in exactly the configurations where
$\sigma_{\mathrm{case}}$ is near zero, so it cannot be used on its own as a
training diagnostic. Second, the downstream metrics vary only modestly across
configurations, because $L_{\mathrm{rec}}$ and the design probe act on the
geometry representation $g$ while the collapse is measured in the target
tokens $Z_t$. Case-level variation therefore has to be monitored directly.

\begin{table}[h]
\centering
\caption{Complete representation-collapse diagnostics. The case-level
standard deviation measures between-case variation, while
$L_{\mathrm{lat}}$ and $L_{\mathrm{rec}}$ report latent prediction and
physical reconstruction losses. Pressure $R^2$ and design-recovery
$R^2$ quantify downstream performance.}
\label{tab:collapse_full}
\small
\setlength{\tabcolsep}{4pt}
\resizebox{\textwidth}{!}{%
\begin{tabular}{llccccc}
\toprule
\textbf{Dataset}
&
\textbf{Configuration}
&
$\sigma_{\mathrm{case}}$
&
$L_{\mathrm{lat}}$
&
$L_{\mathrm{rec}}$
&
Pressure $R^2$
&
Design $R^2$
\\
\midrule
SHIFT-Wing
& None
& 0.0003 & 0.0000 & 0.0354 & 0.9906 & 0.992 \\
& Token SIGReg
& 0.0003 & 0.0000 & 0.0352 & 0.9906 & 0.992 \\
& Case SIGReg
& 0.6211 & 0.0403 & 0.0373 & 0.9893 & 0.987 \\
& + Variance
& 1.0346 & 0.0498 & 0.0381 & 0.9886 & 0.986 \\
& + Target Recon.
& 1.0317 & 0.0447 & 0.0379 & 0.9971 & 0.998 \\
\midrule
SHIFT-Submarine
& None
& 0.0004 & 0.0000 & 0.0348 & 0.9896 & 0.982 \\
& Token SIGReg
& 0.0005 & 0.0000 & 0.0352 & 0.9897 & 0.982 \\
& Case SIGReg
& 0.7566 & 0.0221 & 0.0361 & 0.9892 & 0.980 \\
& + Variance
& 1.0633 & 0.0221 & 0.0360 & 0.9894 & 0.982 \\
& + Target Recon.
& 1.0620 & 0.0192 & 0.0362 & 0.9967 & 0.998 \\
\midrule
SHIFT-SUV
& None
& 0.0002 & 0.0000 & 0.0537 & 0.9726 & 0.975 \\
& Token SIGReg
& 0.0002 & 0.0000 & 0.0534 & 0.9729 & 0.979 \\
& Case SIGReg
& 0.0003 & 0.0000 & 0.0529 & 0.9734 & 0.982 \\
& + Variance
& 0.9892 & 0.1493 & 0.0547 & 0.9718 & 0.967 \\
& + Target Recon.
& 0.9889 & 0.1526 & 0.0551 & 0.9937 & 0.989 \\
\bottomrule
\end{tabular}}
\end{table}

These results motivate the use of case-level constraints in the final
D-JEPA training objective. They also illustrate why token-level diversity
alone is insufficient as a diagnostic for a geometry representation:
variation must be evaluated across geometry cases.

\section{Evaluation Protocols}

\begin{figure}[h]
\centering
\includegraphics[width=\textwidth]{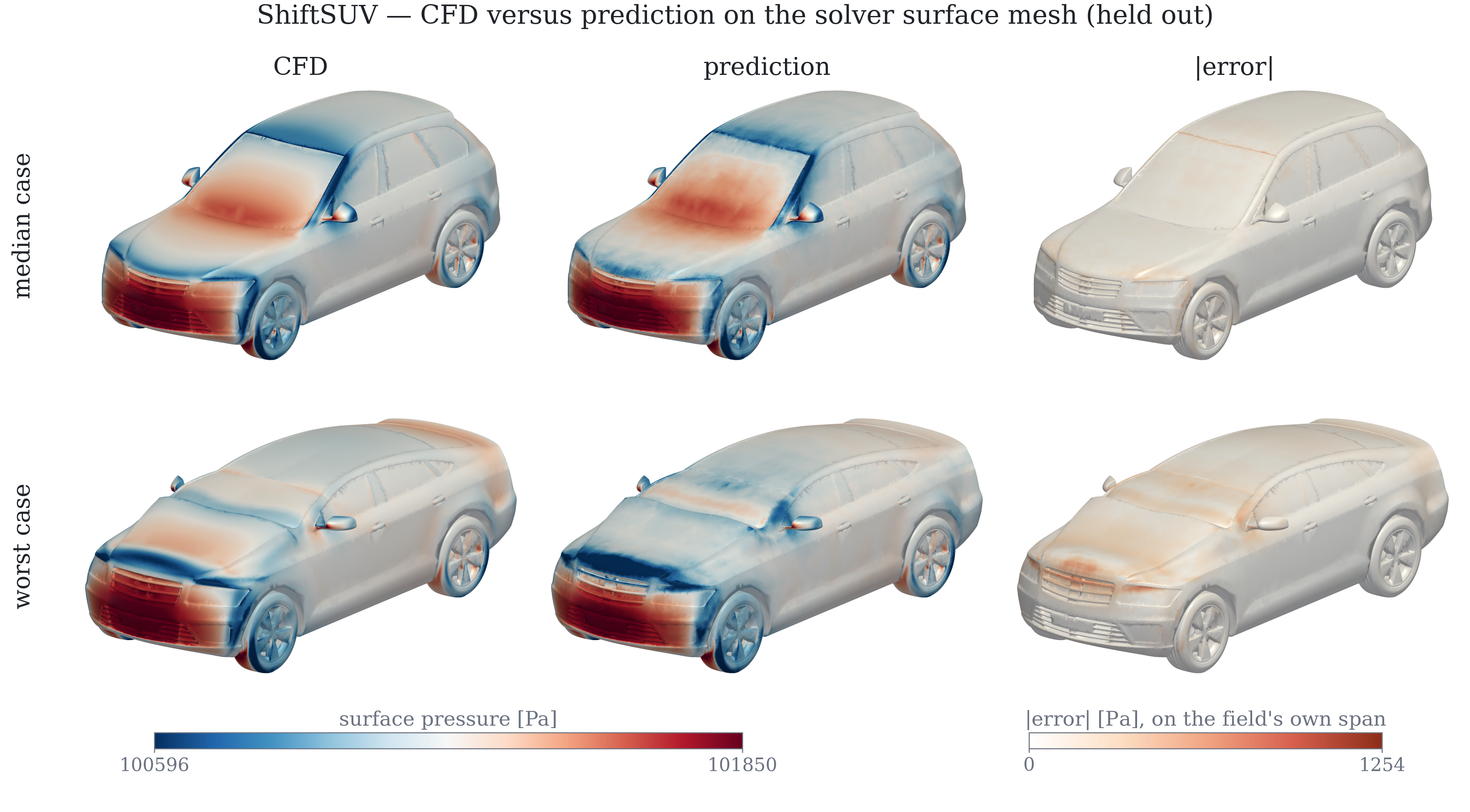}
\caption{Qualitative surface pressure prediction for SHIFT-SUV.
Predicted pressure and absolute point-wise error are shown for
representative median and worst-case held-out geometries.}
\label{fig:a4}
\end{figure}

\begin{figure}[h]
\centering
\includegraphics[width=\textwidth]{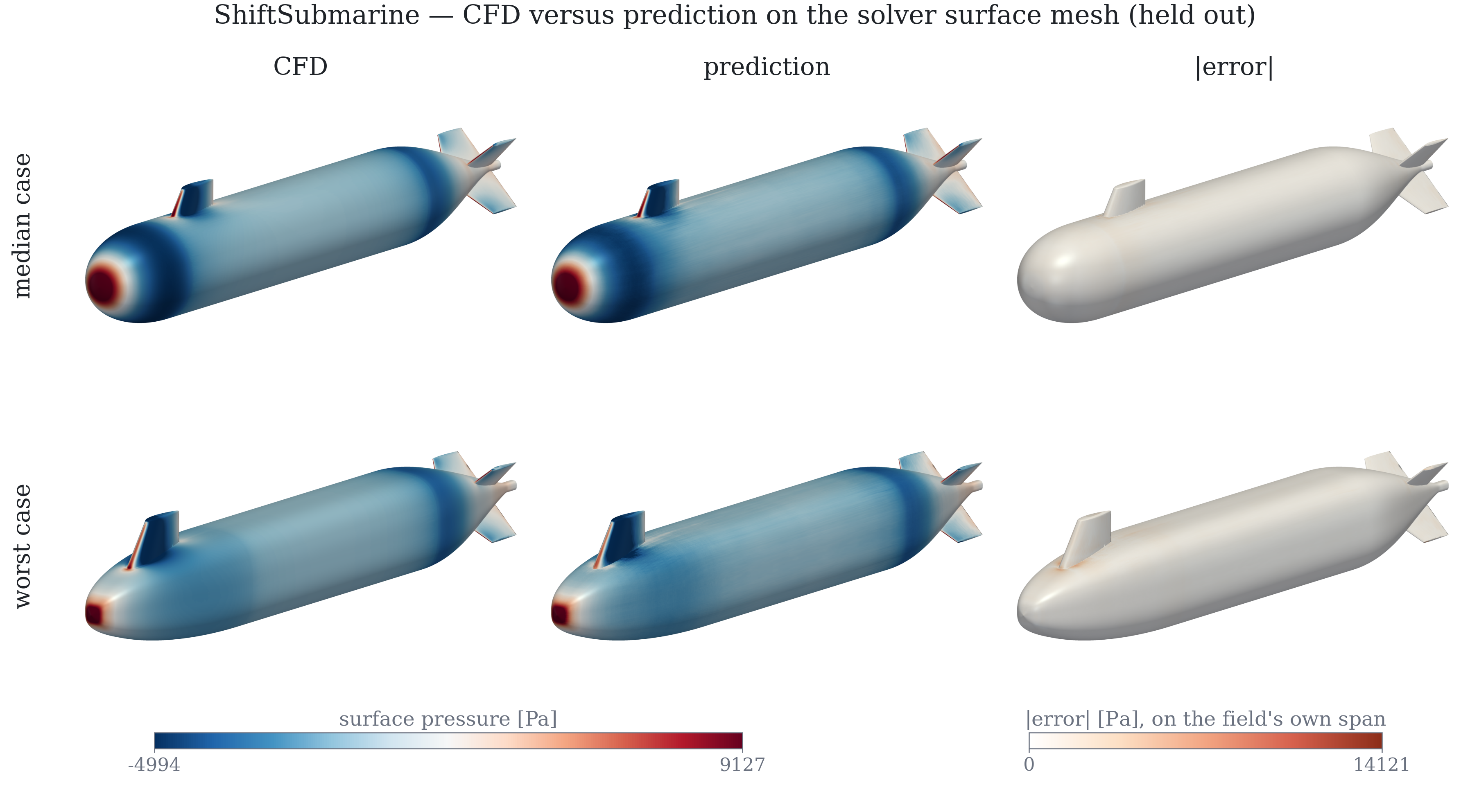}
\caption{Qualitative surface pressure prediction for SHIFT-Submarine.
Predicted pressure and absolute point-wise error are shown for
representative median and worst-case held-out geometries.}
\label{fig:a5}
\end{figure}

\begin{figure}[h]
\centering
\includegraphics[width=\textwidth]{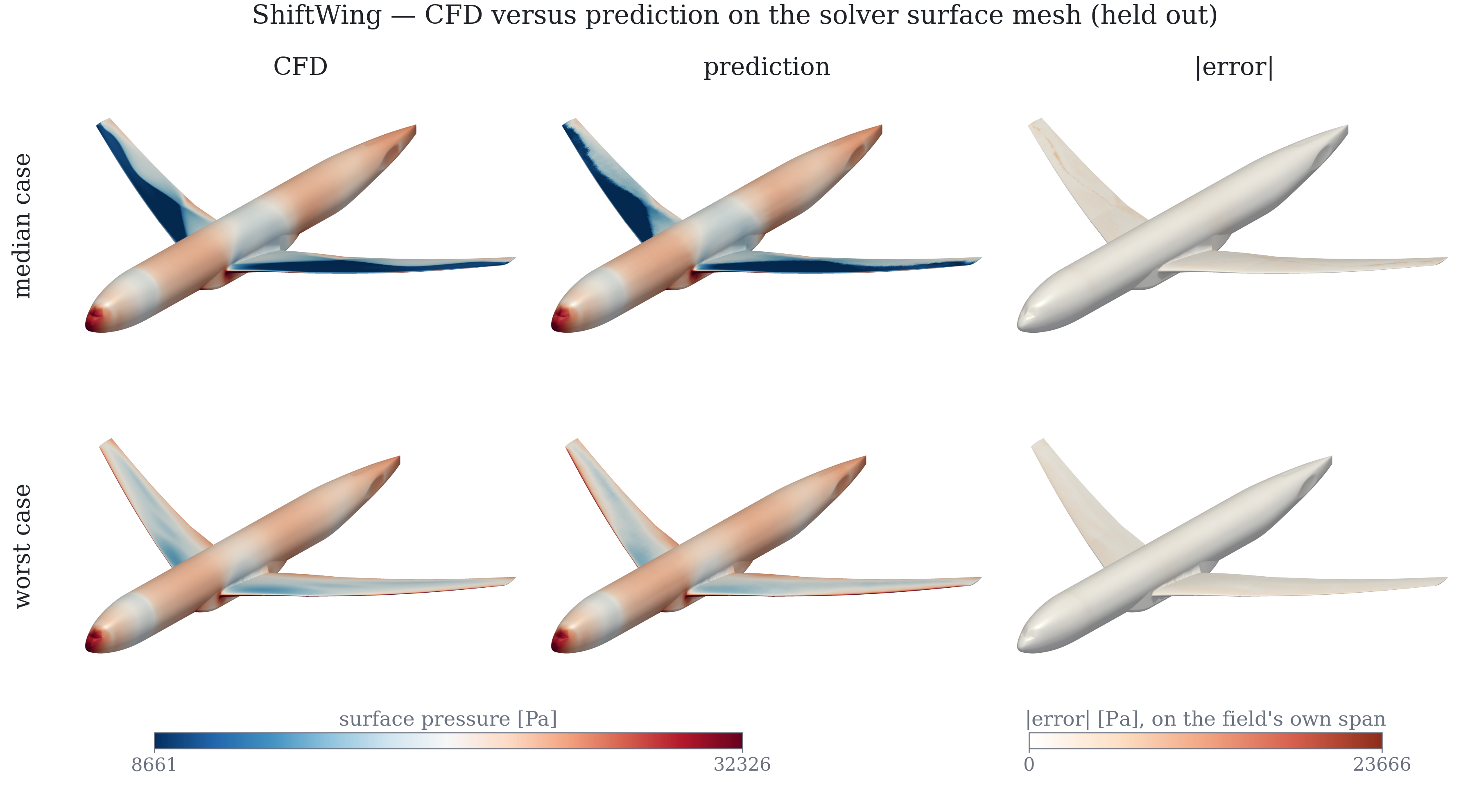}
\caption{Qualitative surface pressure prediction for SHIFT-Wing.
Predicted pressure and absolute point-wise error are shown for
representative median and worst-case held-out geometries.}
\label{fig:a6}
\end{figure}

All benchmark evaluations use fixed 10\% validation partitions generated
with constant random seeding. Validation metrics are evaluated on held-out
geometry cases that are not used for model training. Field prediction accuracy is quantified using the coefficient of
determination $R^2$, computed independently for each physical output
variable and averaged across held-out cases. Reported uncertainty values
are standard deviations across three independent training seeds. Integrated force coefficients, including drag $C_D$ and lift $C_L$, are
computed by direct integration of predicted pressure and wall-shear-stress
fields over the complete solver surface. No secondary regression head is
used to predict the integrated quantities. Design-variable recoverability is evaluated using a standardized,
five-fold cross-validated ridge-regression probe applied to frozen
128-dimensional geometry representations $g$ \citep{alain2017probes}.
Out-of-fold $R^2$ values
are reported to avoid evaluating the probe on the same samples used for
fitting. Qualitative field comparisons are performed on dense, high-resolution
solver meshes rather than only on the sparse point samples used during
training. Representative median and worst-case held-out examples are
shown in Figures~\ref{fig:a4}, \ref{fig:a5} and \ref{fig:a6}.

\section{Baseline Models and Comparative Benchmarks}
\label{app:baselines}

\subsection{AeroJEPA baseline}

AeroJEPA \citep{giral2026aerojepa} is used as the primary
representation-learning baseline. The comparison retains the same
geometry backbone components, point-sampling procedure, optimizer settings
and primary reconstruction objective where applicable. This controls for
differences in model capacity and focuses the comparison on the geometry
representation and additional D-JEPA training objectives. For design recoverability comparison, the pooled 128-dimensional
AeroJEPA context representation is evaluated using the same linear probing
procedure applied to D-JEPA.

\subsection{GeoTransolver and DoMINO baselines}

GeoTransolver \citep{adams2025geotransolver} and DoMINO
\citep{ranade2025domino} are evaluated using their official
implementations on the same dataset splits. GeoTransolver uses 20
transformer layers, hidden width 256, eight attention heads, 128 slices
and a 60{,}000-point surface representation under the corresponding
optimization settings.

All baseline metrics are computed on held-out validation cases. The
resulting field-prediction and integrated-force comparisons are reported
in Table~\ref{tab:main} of the main text.

\section{Supplemental Experimental Results}
\label{app:supp}

\subsection{Design-variable recoverability}

A central property of D-JEPA is that the geometry representation retains
information about the underlying design variables. We evaluate this
property independently of the physics decoder by freezing the learned
128-dimensional geometry representation and fitting a linear ridge probe. Table~\ref{tab:recover} reports the recovery accuracy for every design
variable. Mean $R^2$ values are 0.989 for SHIFT-SUV, 0.998 for SHIFT-Wing
and 0.998 for SHIFT-Submarine.

\begin{table}[h]
\centering
\caption{Linear-probe recovery accuracy ($R^2$) for individual geometric
design parameters from the frozen 128-dimensional geometry representation
$g$.}
\label{tab:recover}
\small
\setlength{\tabcolsep}{4pt}
\resizebox{\textwidth}{!}{%
\begin{tabular}{lc@{\hskip 1.2em}lc@{\hskip 1.2em}lc}
\toprule
\multicolumn{2}{c}{\textbf{SHIFT-SUV}} &
\multicolumn{2}{c}{\textbf{SHIFT-Wing}} &
\multicolumn{2}{c}{\textbf{SHIFT-Submarine}} \\
\cmidrule(lr){1-2}
\cmidrule(lr){3-4}
\cmidrule(lr){5-6}
Parameter & $R^2$ &
Parameter & $R^2$ &
Parameter & $R^2$ \\
\midrule
hood\_height
& 0.995
& AR (aspect ratio)
& 0.999
& cross\_section\_shape
& 1.000 \\

hood\_base\_height
& 0.987
& Lambda\_c\_4 (quarter-chord sweep)
& 0.998
& fore\_body\_shape
& 0.996 \\

windshield\_angle
& 0.986
& twist\_20
& 0.997
& fore\_length
& 0.999 \\

vehicle\_height
& 0.991
& twist\_k
& 0.997
& mid\_length
& 0.999 \\

backlight\_angle
& 0.990
& twist\_60
& 0.999
& aft\_tail\_half\_angle
& 0.997 \\

rr\_tapering
& 0.988
& twist\_80
& 0.999
& sail\_length\_ratio
& 0.997 \\

fr\_plainview
& 0.986
& twist\_t
& 0.998
& strake\_length\_ratio
& 0.996 \\

\midrule
\textbf{Mean}
& \textbf{0.989}
& \textbf{Mean}
& \textbf{0.998}
& \textbf{Mean}
& \textbf{0.998} \\
\bottomrule
\end{tabular}}
\end{table}

\begin{figure}[h]
\centering
\includegraphics[width=\textwidth]{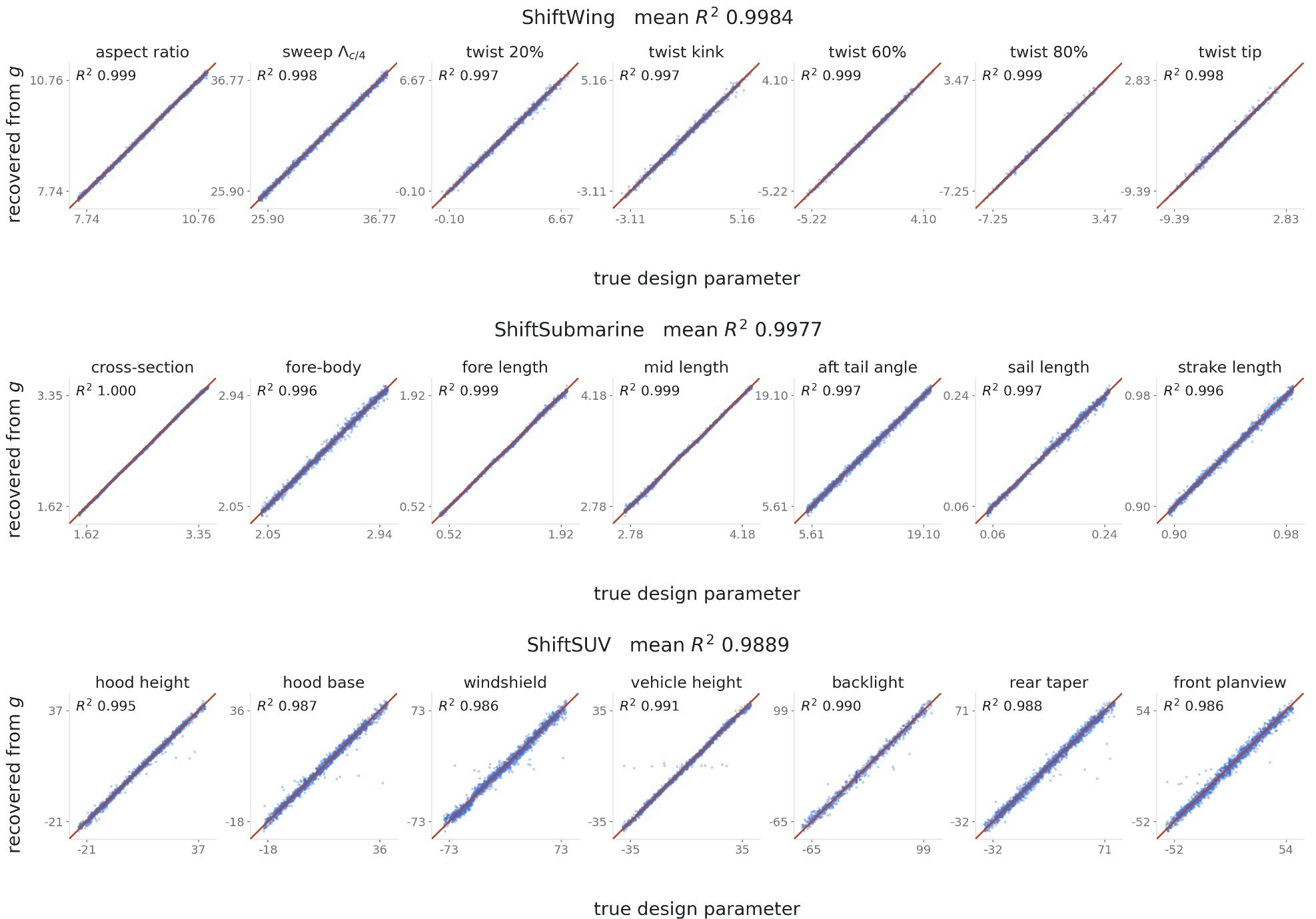}
\caption{Design-variable probe recoverability across benchmark datasets.
True design parameters are compared with values recovered from the frozen
geometry representation using linear probes.}
\label{fig:a7}
\end{figure}

Individual design-parameter $R^2$ values range from 0.986 to 0.995 for
SHIFT-SUV, 0.997 to 0.999 for SHIFT-Wing and 0.996 to 1.000 for
SHIFT-Submarine. These results indicate that the learned geometry
representation retains substantial linearly recoverable information about
the underlying parameterization. Figure~\ref{fig:a7} shows the
corresponding per-parameter probe correlations.

\subsection{Continuous operating-condition generalization}

\begin{figure}[h]
\centering
\includegraphics[width=\textwidth]{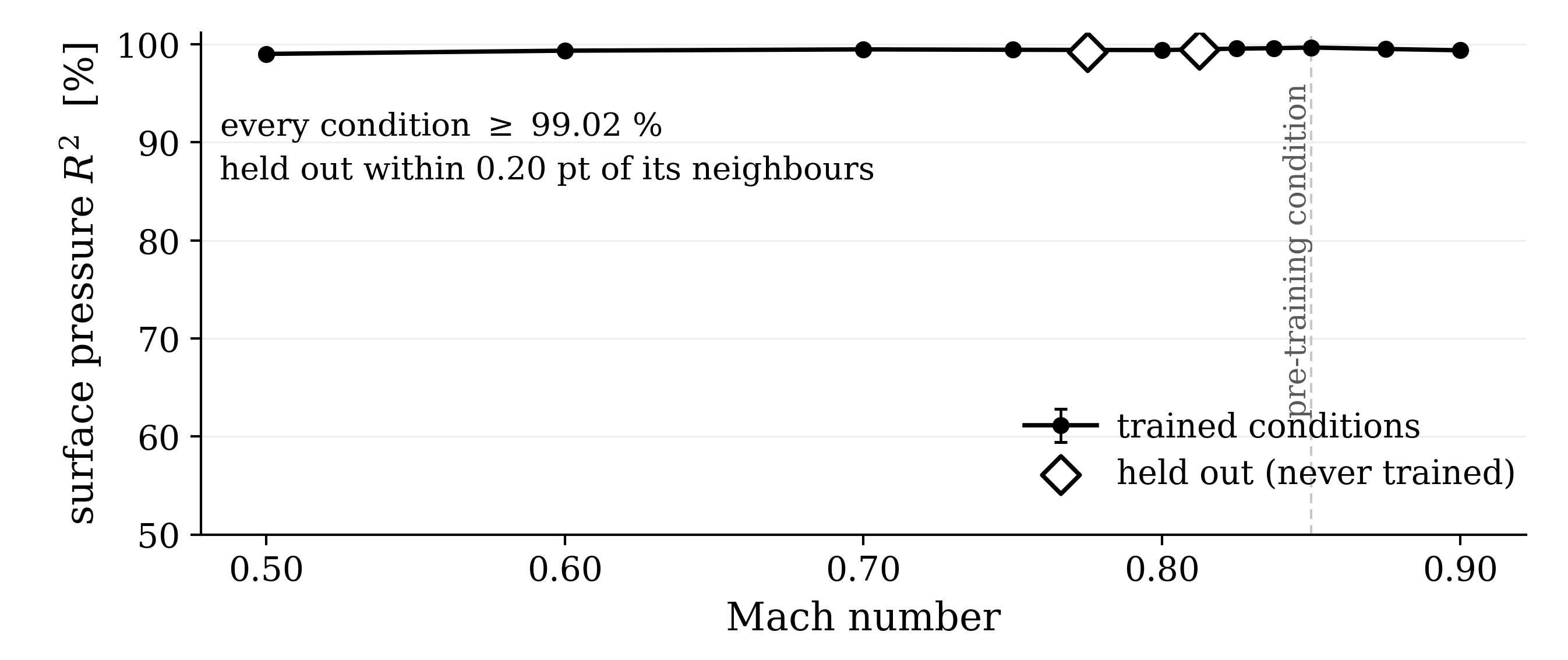}
\caption{Operating-condition generalization across Mach numbers on
SHIFT-Wing. Surface pressure $R^2$ is evaluated across trained and
held-out Mach conditions using the same geometry representation.}
\label{fig:a8}
\end{figure}

Because the geometry representation does not receive the operating
condition vector, the same geometry representation can be reused across
different flow conditions while the decoder conditions the field prediction
on $c$. On SHIFT-Wing, the evaluated Mach range is 0.50--0.90. Ten Mach states are
used during training, while Mach 0.7750 and Mach 0.8125 are held out
entirely. Pressure prediction remains above $R^2=0.990$ across the
evaluated conditions, with $R^2=0.9922$ and $R^2=0.9944$ at the two held-out
Mach numbers. Figure~\ref{fig:a8} shows the accuracy across the full sweep.

\subsection{Cross-physics decoder transfer}

To evaluate reuse across physical response spaces, a pre-trained
aerodynamic geometry encoder is frozen and transferred to AutoHood 3D
structural displacement prediction, following the frozen-backbone strategy
common in parameter-efficient adaptation \citep{houlsby2019adapters,hu2022lora}.
The geometry encoder and its projection
layers remain fixed, while only the new physics decoder is trained. The
transfer configuration contains 0.56\,M trainable decoder parameters,
compared with 2.30\,M for a model trained entirely from scratch, a
$4.1\times$ reduction. Per-epoch results are reported in
Table~\ref{tab:crossphysics} of the main text.

\section{Differentiable Design Optimization}
\label{app:opt}

D-JEPA supports differentiable design optimization because gradients
can be propagated from a design objective through the physics decoder and
learned geometry representation back to the underlying design variables.
Specifically, for design variables $\theta$, the differentiable path is
\[
\theta \rightarrow G(\theta) \rightarrow (P,N)
\rightarrow E_\vartheta(P,N)=g
\rightarrow D_{\phi_s}(g,q,c)
\rightarrow \hat{F}
\rightarrow J(\hat{F}),
\]
where $G(\theta)$ generates the corresponding geometry and $J$ denotes the
design objective computed from the predicted physical field. We evaluate
this capability as a downstream application of the learned surrogate rather
than as a separate representation-learning objective. Gradients are computed
using automatic differentiation through the geometry representation and
physics decoder. For SHIFT-Wing,
SHIFT-SUV and SHIFT-Submarine, optimization trajectories show strong
alignment between the optimized objective and latent search directions,
with objective correlation values of approximately $R^2=0.99$, $0.99$ and
$0.98$, respectively. To keep optimization within the learned design space,
design-variable bounds are imposed through linear design probes together
with a $k$-nearest-neighbor trust-region constraint. The candidate designs
produced by this procedure are the ones validated against high-fidelity CFD
in Table~\ref{tab:cfd}.

\section{Computational Hardware and Runtime Requirements}

The primary hardware configurations and approximate wall-clock runtimes
are summarized in Table~\ref{tab:a7}. Runtime values are provided to
contextualize the computational requirements of the reported experiments. For decoder-transfer experiments, freezing the geometry encoder reduces the
number of trainable parameters and eliminates backward-gradient storage
through the frozen representation. A forward pass through the frozen
encoder is still required. Consequently, transfer efficiency arises from
both reduced optimizer memory requirements and the removal of backward
computation through the geometry encoder.

\begin{table}[h]
\centering
\caption{Primary computational hardware configurations and approximate
wall-clock runtimes across pipeline stages.}
\label{tab:a7}
\small
\begin{tabular}{lll}
\toprule
\textbf{Execution pipeline stage}
&
\textbf{Compute hardware}
&
\textbf{Wall-clock time}
\\
\midrule
D-JEPA/AeroJEPA training (400 epochs)
&
1 $\times$ NVIDIA T4 (16\,GB)
&
4--8 hours / dataset
\\

Decoder transfer fine-tuning (400 epochs)
&
1 $\times$ NVIDIA T4 (16\,GB)
&
$\sim$2 hours
\\

GeoTransolver baseline (500 epochs)
&
8 $\times$ NVIDIA H100 (80\,GB)
&
$\sim$12 hours / dataset
\\

Mach-condition fine-tuning (200 epochs)
&
1 $\times$ NVIDIA T4 (16\,GB)
&
$\sim$3 hours
\\
\bottomrule
\end{tabular}
\end{table}

The reported runtimes depend on dataset size, solver mesh resolution,
batching and hardware utilization and should therefore be interpreted as
approximate execution measurements rather than hardware-independent
complexity estimates.

\section{Limitations}

The current evaluation is limited to a small set of parameterized engineering
datasets, primarily covering aerodynamic and structural response fields, with
cross-physics transfer demonstrated in a single CFD-to-structural setting.
Although the results show strong design-variable recoverability, the degree of
recoverability varies across geometry families, suggesting that some geometric
degrees of freedom may be less explicitly represented in the learned latent
space. The operating-condition transfer experiments are also evaluated over a
limited range of conditions and do not establish robustness to substantially
different flow regimes or boundary conditions. Similarly, the optimization
experiments use differentiable surrogate models and predefined objectives
within the evaluated design spaces, whereas practical engineering problems may
involve constraints, discrete variables, noisy objectives and stronger
out-of-distribution conditions. Broader multi-physics benchmarks, more diverse
geometry families and constrained real-world optimization therefore remain
important directions for future evaluation.
\end{document}